\pdfoutput=1

\documentclass[11pt]{article}

\usepackage[final]{acl}

\usepackage{times}
\usepackage{latexsym}
\usepackage{adjustbox}
\usepackage[most]{tcolorbox}
\usepackage{hyperref}
\usepackage{booktabs}
\usepackage{multirow} 
\usepackage{colortbl}
\usepackage{subcaption}

\usepackage[T1]{fontenc}

\usepackage[utf8]{inputenc}

\usepackage{microtype}
\usepackage{amsmath}
\usepackage{amssymb}
\usepackage{algorithm}
\usepackage{algorithmic}

\usepackage{inconsolata}

\usepackage{graphicx}

\title{Towards Open-Ended Emotional Support Conversations in LLMs via Reinforcement Learning with Future-Oriented Rewards}

\author{
 \textbf{Ting Yang\textsuperscript{1}},
 \textbf{Li Chen\textsuperscript{1}},
 \textbf{Huimin Wang\textsuperscript{2}}
\\
\\
 \textsuperscript{1}Hong Kong Baptist University, Hong Kong, China \\
 \textsuperscript{2}Jarvis Lab, Tencent, Shenzhen, China
\\
\{{cstyang,lichen}\}@comp.hkbu.edu.hk \\
  hmmmwang@tencent.com
 }

\begin{document}
\maketitle
\begin{abstract}
Emotional Support Conversation (ESC) systems aim to alleviate users' emotional difficulties and provide long-term, systematic support for emotional well-being. However, most large language model (LLM)-based ESC systems rely on predefined strategies, which limits their effectiveness in complex, real-life scenarios. To enable flexible responses to diverse emotional problem scenarios, this paper introduces a novel end-to-end framework (RLFF-ESC) that directly learns enduring emotionally supportive response skills using reinforcement learning. For sustained emotional support, we first employ an LLM-based multi-agent mechanism to simulate future dialogue trajectories and collect future-oriented rewards. We then train a future-oriented reward model, which is subsequently used to train the emotional support policy model. Additionally, we incorporate an explicit reasoning process during response generation to further enhance the quality, relevance, and contextual appropriateness of the system's responses. We evaluate the backbone policy model on Qwen2.5-7B-Instruct-1M and LLaMA3.1-8B-Instruct models, testing the proposed RLFF-ESC framework across two public ESC datasets. Experimental results demonstrate that RLFF-ESC consistently outperforms existing baselines in terms of goal completion and response quality.
\end{abstract}

\section{Introduction}

\begin{figure}[h]
    \centering
    \includegraphics[width=\linewidth]{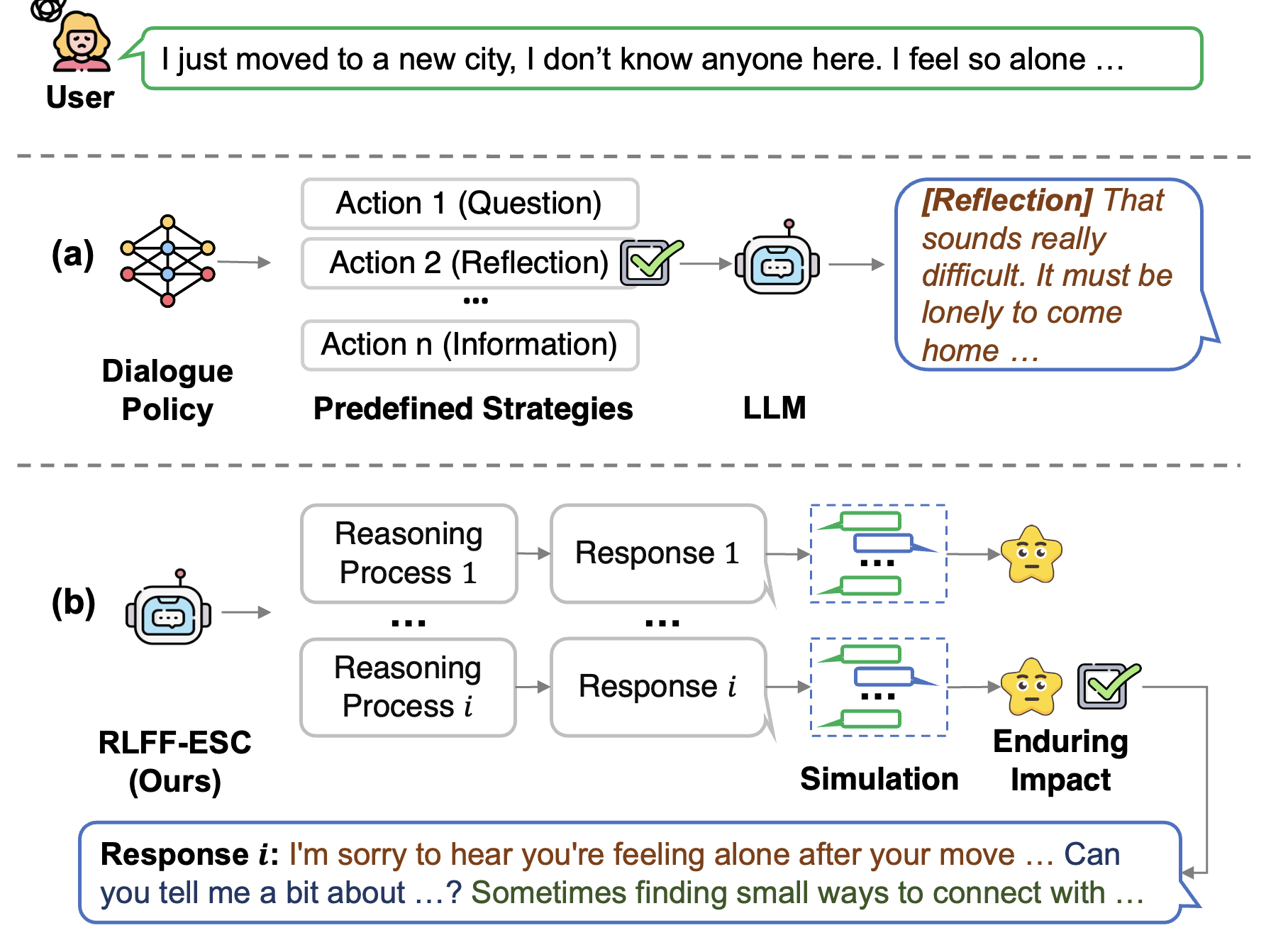}
     \caption{Comparison of (a) traditional strategy planning methods, which select predefined supportive strategies before generating responses, and (b) our approach, which directly reasons from conversation context to generate open-ended and emotionally supportive responses by simulating the long-term emotional impact of future dialogue.}
    \label{fig:intro}
\end{figure}

Online emotional support conversation systems~\cite{burleson:2003, heaney:2008} aim to alleviate people's emotional distress and promote psychological well-being by offering interactive communication and active listening through digital platforms. 
The rapid development of large language models~\cite{chang_llm:2024}, such as GPT-4~\cite{achiam:2023} and LLaMA~\cite{grattafiori:2024}, has sparked growing interest in leveraging these models to facilitate high-quality emotional support conversations, given their dedicated contextual understanding and response generation capabilities~\cite{yi_dial:2024}.
Recent research has explored improving ESC using LLMs through techniques such as prompt engineering~\cite{chen:2023, zhao:2023, zhang_escot:2024}, and supervised fine-tuning~\cite{zheng:2023, chen:2025}. 
However, these approaches primarily emphasize immediate response quality, which is assessed by how well the generated responses align with ground truth, while overlooking the long-term effects that such responses may have on the user's emotional states over the course of the conversation.

Effective emotional support requires consideration of both immediate and long-term effects of supportive efforts~\cite{burleson:2003}. It aims not only to alleviate present distress but also to guide users toward sustained emotional well-being~\cite{park:2023}.
To this end, recent studies have incorporated long-term considerations into the strategy planning of LLMs~\cite{deng:2024, he:2024, zhao:2025, wang:2025}, often by integrating them with external strategy planners to guide response generation.
However, these approaches typically rely on predefined strategy sets (e.g., \textit{Question}, \textit{Reflection of Feelings}), thereby limiting the LLM’s behavior to a constrained strategy space.  As a result, the flexibility of these systems remain limited, especially in dynamic and nuanced emotional support contexts~\cite{burleson:2013}.
Moreover, they primarily focus on optimizing external strategy planners, while underutilizing the intrinsic reasoning capabilities of LLMs (e.g., reasoning about users’ underlying emotions and distress cues), which are critical for providing effective emotional support.

To address the aforementioned limitations, we propose \underline{\textbf{R}}einforcement \underline{\textbf{L}}earning from \underline{\textbf{F}}uture-oriented \underline{\textbf{F}}eedback for \underline{\textbf{E}}motional \underline{\textbf{S}}upport \underline{\textbf{C}}onversations (\textbf{RLFF-ESC}),  
a new end-to-end framework that directly optimizes LLMs using reinforcement learning. 
RLFF-ESC leverages future-oriented feedback, specifically, the anticipated emotional responses of users over time, as reward signals to guide the optimization of LLMs. This encourages the model to generate emotionally supportive responses that are conducive to enduring emotional improvement.
Unlike previous methods, RLFF-ESC emphasizes the LLM's self-reflective reasoning in complex emotional support scenarios, enabling more open-ended and adaptive response generation (as shown in Figure~\ref{fig:intro}).
Our central research question is: \textit{How can LLMs be optimized to generate open-ended, emotionally supportive responses that account for enduring emotional outcomes?}

Specifically, rather than restricting responses to predefined strategies, we prompt LLMs to generate intermediate reasoning processes, such as understanding the user’s emotional state, identifying cues of emotional distress, and formulating adaptive support strategies tailored to the specific emotional context.
During training, the LLM generates multiple candidate outputs, each comprising an explicit reasoning process followed by a supportive response. To evaluate these candidates, we design a custom reward function with two components: a \textit{thinking format reward}, which enforces adherence to a structured reasoning–response format, and a \textit{future-oriented reward} derived from a learned neural reward model that estimates the enduring emotional impact of the generated response. 
We then optimize the LLM policy by using Group Relative Policy Optimization (GRPO)~\cite{shao:2024}, a reinforcement learning algorithm that enables stable and efficient policy updates through group-based computations, guided by the designed reward signals.

To estimate the future-oriented rewards of responses generated by the LLM-based ESC system, we propose a multi-agent dialogue simulation module. Given that the emotional outcomes of a system response can accumulate over multiple turns, depending on the user's evolving emotional state, the module samples future dialogue trajectories between a user simulator and the LLM-based ESC system.
To evaluate these simulated dialogue trajectories, we adopt an LLM-based critic model, motivated by recent findings showing that LLMs can effectively approximate human judgments in subjective evaluation tasks~\cite{gu:2024}. This eliminates the need for costly human annotation, enabling a scalable and cost-effective solution for assessing emotional impact.
The critic determines whether the user’s emotional issue has been resolved and assigns an estimated reward accordingly. The simulation continues until either the emotional issue is resolved or a predefined maximum number of dialogue turns is reached, at which point the future-oriented reward for the given system response is obtained.

We validate the effectiveness of our proposed framework, RLFF-ESC, through comprehensive experiments conducted on two public datasets. 
The main contributions of this work are as follows: 
\begin{itemize}    
    \item We propose RLFF-ESC, a novel end-to-end framework that uses reinforcement learning from future-oriented feedback to guide large LLMs in reasoning over dynamic emotional contexts and generating open-ended responses aligned with enduring emotional outcomes.
    \item We design a multi-agent dialogue simulation module to estimate future-oriented rewards and develop an effective reward function for the LLM policy optimization via GRPO.
    \item We conduct extensive experiments, which demonstrate that RLFF-ESC  outperforms existing baseline methods in task completion and high-quality response generation.
\end{itemize}

\section{Related Work}
\citet{liu:2021} introduce the emotional support conversation task and release the ESConv dataset for its development.
In the pre-LLM era, researchers have explored various methods to model the users' emotional states \cite{peng:2022, cheng:2023, jia:2023, deng:2023} and the strategy learning \cite{cheng:2022, tu:2022} in ESC systems.
For example, \citet{peng:2022} propose a hierarchical graph network to model emotional causes and users' psychological intentions. \citet{cheng:2023} infer user personas from conversation histories to provide more personalized emotional support. 
Furthermore, external knowledge has been incorporated to improve understanding of users' emotional distresses~\cite{jia:2023, deng:2023}.
To improve strategy learning, \citet{cheng:2022} employ a look-ahead heuristic based on the A* search algorithm, \citet{tu:2022} model strategies as probability distributions to generate mixed-strategy responses.

With the rise of LLMs, recent studies have explored their application in ESC by prompting and supervised fine-tuning \cite{chen:2023, zhao:2023, qiu:2024, chen:2025}. 
For example, \citet{zhang_escot:2024} apply Chain-of-Thought (CoT) to improve response interpretability. 
\citet{zheng:2023} utilize ChatGPT to synthesize the ExTES dataset and fine-tune a LLaMA model for ESC.
Other works integrate LLMs with external policy model to improve strategy selection~\cite{wan:2025, zhang:2023}. 
For instance, \citet{kang:2024} show that using external planners mitigates LLMs' strategy preference bias.
To account for the long-term impact of supportive strategies~\cite{he:2024}, \citet{yu:2023} prompt LLMs to perform Monte Carlo Tree Search (MCTS) for goal-oriented policy planning, while \citet{fu:2023} use self-play simulations with critique-based feedback to assess long-term effects. 
\citet{deng:2024} apply reinforcement learning with cumulative rewards to train a plug-in dialogue planner for strategy prediction. 
DialogueXpert~\cite{rakib:2025} uses a frozen LLM as an action prior, along with an emotion tracker, to enable lightweight Q-network training for optimal action selection.

Reinforcement learning has recently emerged as an effective technique for aligning LLM responses with human preferences.
One of the most widely adopted approaches is Reinforcement Learning from Human Feedback (RLHF), which trains a preference model on human-annotated data and uses Proximal Policy Optimization (PPO) to fine-tune the LLM based on preference scores \cite{bai:2022}. 
To simplify preference learning, \citet{rafailov:2023} propose Direct Preference Optimization (DPO), which bypasses reward modeling by directly optimizing the LLM policy to align human preferences. 
Group Relative Policy Optimization~\cite{shao:2024} removes the traditional critic model and introduces a group-based evaluation strategy. 
RL-based LLM optimization has recently been adapted for ESC to better align LLM outputs with users' emotional needs.
For example, \citet{zhao:2025} construct a strategy-level conversation tree via Monte Carlo Tree Search and employ DPO to fine-tune LLMs, enhancing their ability to select appropriate support strategies.
In contrast to prior methods that focus primarily on optimizing strategy selection, our approach directly optimizes the LLM with GRPO using future-oriented rewards to promote self-reflective reasoning in dynamic emotional support scenarios, thereby enabling open-ended responses that account for enduring emotional impact.

\begin{figure*}[ht!]
  \centering
    \begin{adjustbox}{width=\textwidth}
    \includegraphics{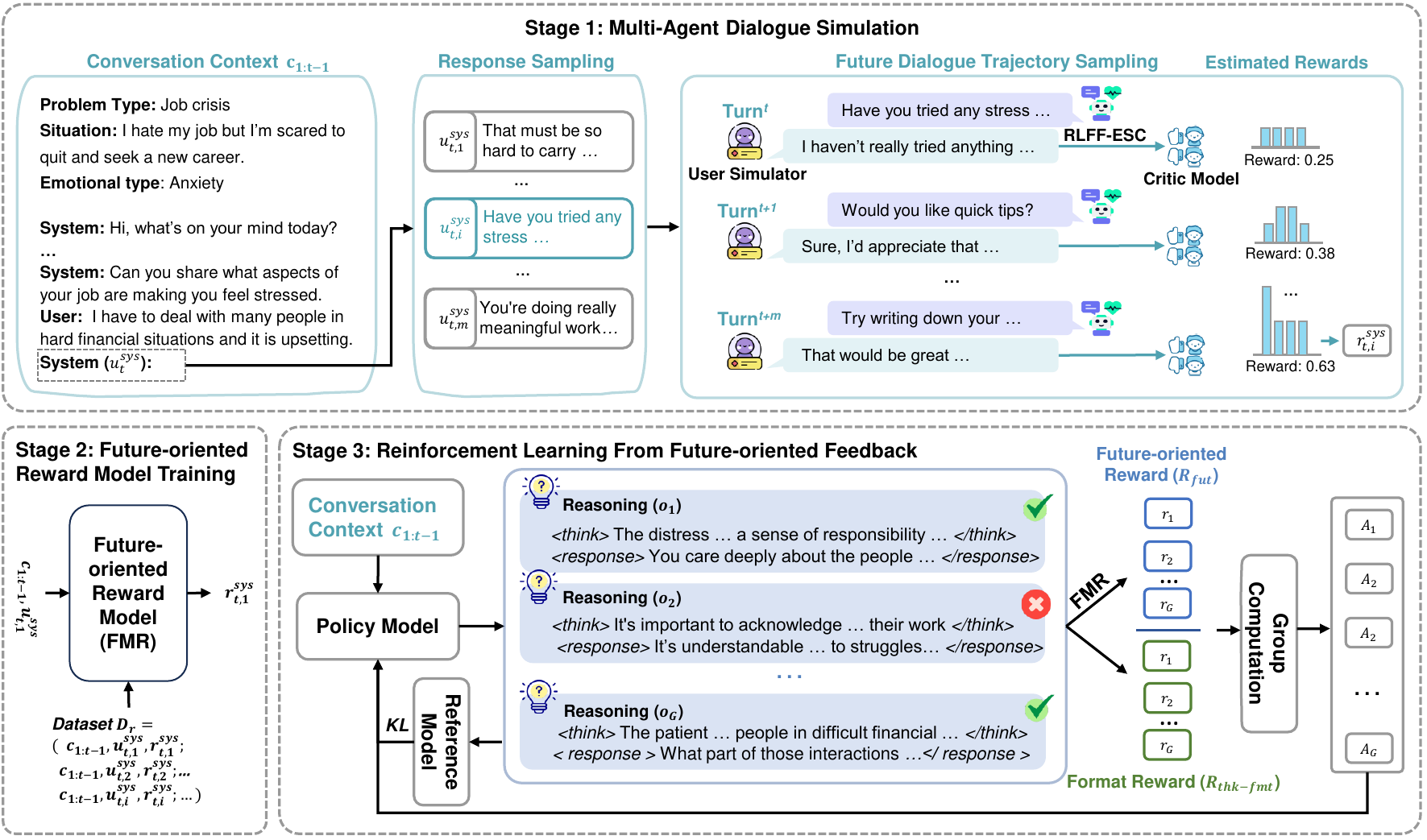}
  \end{adjustbox}
  \caption{Overview of the RLFF-ESC framework. \textbf{Stage 1:} Multi-agent dialogue simulation samples future dialogue trajectories to estimate rewards for responses. \textbf{Stage 2:} A reward function is designed to guide RL, including training a reward model for future-oriented reward prediction. \textbf{Stage 3:} Reinforcement learning with GRPO is employed to optimize the LLM, encouraging the generation of open-ended responses that consider enduring emotional impact.}
  \label{framework}
\end{figure*}

\section{Methodology}
The RLFF-ESC framework, as illustrated in Figure \ref{framework}, consists of three stages: multi-agent dialogue simulation, future-oriented reward model training, and reinforcement learning from future-oriented feedback.

\subsection{Problem Formulation}
Given a conversational context comprising dialogue history and a scenario description (e.g., problem type, emotion type, and situational details), it can be represented as \( c_{1:t-1} = \{meta\_des, u_1^{sys}, u_1^{usr}, \dots, u_{t-1}^{sys}, u_{t-1}^{usr}\} \sim \mathcal{D} \), where \(\mathcal{D}\) represents the dataset, \(meta\_des\) is the scenario description and each utterance \(u_i = \{w_1^i, w_2^i, \dots, w_n^i\}\) is a sequence of \(n\) words. The target is to generate the next system response \(u_t^{sys}\) that is coherent with the conversation context and effective in achieving a specific objective, such as reducing the user's emotional distress. Formally, the goal is to learn a policy \(\pi_\theta\) that generates system responses to maximize the expected reward over the observed dialogue:
\begin{equation}
   \pi^* = \arg\max_{\pi_\theta} \mathbb{E}_{x \sim \mathcal{D}, \, y \sim \pi_\theta(\cdot \,|\, x)} \big[ R(x, y) \big],
\end{equation}
where \( x = c_{1:t-1} \), \( y = u_{t}^{sys} \), and \( R(\cdot) \) is a function that estimates the rewards of the system’s responses.

\subsection{Multi-Agent Dialogue Simulation}
To estimate the enduring impact of a system response, we employ a multi-agent simulation framework that generates future dialogue trajectories. Given the conversation context \( c_{1:t-1} \), we first sample a set of candidate next-turn responses \( \{ u_{t,j}^{sys} \}_{j=1}^{m} \) from the current policy of the LLM-based ESC model \( M_{sys} \), such that \( u_{t,j}^{sys} \sim P_{M_{sys}}(\cdot \mid p_{sys}, c_{1:t-1}) \),  where \(p_{sys}\) is the prompt. Our target is to construct a dataset consisting of pairs of sampled responses and their corresponding rewards, which can then be used to train a reward model \( R(\cdot) \) for future-oriented reward estimation.

To evaluate the future-oriented reward of each sampled response, we simulate the potential future trajectory of the dialogue. This simulation involves three LLM-based agents: a \textit{system agent} \(M_{sys}\), a \textit{user simulator} \(U\), and a \textit{critic agent} \(M_{crt}\). 
The system agent interacts with the user simulator by generating system responses based on the ongoing dialogue. 
The user simulator emulates a user seeking emotional support, responding based on the conversation history. 
Meanwhile, the critic agent evaluates the conversation's progress according to predefined criteria, such as goal achievement. This simulation-based framework allows for the estimation of future-oriented rewards by capturing how individual responses influence the overall success and quality of the conversation over time.

The future dialogue trajectory simulated for each sampled response is denoted as \(C_{t,j}^{f} = (u_{t,j}^{usr}, \dots, u_{t+k,j}^{sys}, u_{t+k,j}^{usr}, \dots, u_{T,j}^{sys}, u_{T,j}^{usr})\), where \( u_{t+k,j}^{usr} \) is generated by the user simulator, and \( u_{t+k,j}^{sys} \) is generated by \(M_{sys}\). At each step \( k \), the critic model \( M_{crt} \) evaluates a reward based on the ongoing simulated conversation. The simulation proceeds iteratively until either the emotional support objective is achieved or a predefined maximum number of dialogue turns \( T \) is reached.

To compute the future-oriented reward \( r_{t,j}^{sys} \) for the sampled system response \( u_{t,j}^{sys} \), we consider both the reward at the terminal step \( T \) and the average number of turns required to achieve the dialogue goal. The resulting dataset 
\(\mathcal{D}_r = \{ (c_{1:t-1}, u_{t,j}^{sys}, r_{t,j}^{sys}) \}_{t,j}\) is then used to train a reward model \( R(\cdot) \), which predicts the future-oriented reward of system responses. Subsequently, this reward model guides the optimization of the system policy \( \pi_\theta \).
The complete simulation procedure is detailed in Algorithm~\ref{alg:simulation}.

\begin{algorithm}[t]
\small
\caption{Multi-Agent Dialogue Simulation for Future-oriented Reward Estimation}
\label{alg:simulation}
\begin{algorithmic}[1]
\REQUIRE Dialogue dataset \( \mathcal{D} = \{ c\}_{i=1}^{Z} \); ESC model \( M_{sys} \); user simulator \( U \); critic model \( M_{crt} \); prompt \(p_{sys}\), \(p_{usr}\), \(p_{crt}\); number of sampled responses \( m \); maximum dialogue turns \( T \)
\ENSURE Simulated dataset \( \mathcal{D}_r = \left\{\left\{\left(c_{1:t}, u_{t,j}^{sys}, r_{t,j}^{sys} \right) \right\}_{j=1}^m \right\}_{i=1}^{Z} \)

\FOR{each conversation context \( c_{1:t-1} \) from \( \mathcal{D} \)}
    \FOR{each \( j = 1 \) to \( m \)}
        \STATE Sample \( u_{t,j}^{sys} \sim P_{M_{sys}}(\cdot \mid p_{sys}, c_{1:t-1}) \)
        \STATE Generate \( u_{t,j}^{usr} \leftarrow P_{U}(p_{usr}, [c_{1:t-1}, u_{t,j}^{sys}]) \)
        \STATE Initialize \( C_{t,j}^{f} \leftarrow [u_{t,j}^{usr}] \)
        \FOR{each step \( k = 1 \) to \( T \)}
            \STATE Generate \( u_{t+k,j}^{sys} \leftarrow P_{M_{sys}}(p_{sys}, [c_{1:t-1}, C_{t,j}^{f}])\)
            \STATE Append \(u_{t+k,j}^{sys}\) to \(C_{t,j}^{f}\)
            \STATE Generate \( u_{t+k,j}^{usr} \leftarrow P_{U}(p_{usr}, [c_{1:t-1}, C_{t,j}^{f}]) \)
            \STATE Append \( u_{t+k,j}^{usr} \) to \( C_{t,j}^{f} \)
            \STATE Evaluate \( r_{t+k,j} \leftarrow P_{M_{crt}}(p_{crt}, [c_{1:t-1}, C_{t,j}^{f}]) \)
            \IF {\(r_{t+k,j} > \text{threshold}\) or \( k = T \)}
                \STATE Compute average turn: \( \text{AvgT} = k + 1 \)
                \STATE Compute reward: \( r_{t,j}^{\text{sys}} = \frac{r_{t+k,j} + \frac{1}{\text{AvgT}}}{2} \)
                \STATE \textbf{break}
            \ENDIF
        \ENDFOR    
        \STATE Add \( (u_{t,j}^{\text{sys}}, r_{t,j}^{\text{sys}}) \) to \( \mathcal{D}_{r} \)
    \ENDFOR
\ENDFOR
\RETURN \( \mathcal{D}_{r} \)
\end{algorithmic}
\end{algorithm}

\subsection{RL from Future-oriented Feedback}
We optimize the LLM using reinforcement learning (RL) to enhance its ability to generate open-ended responses that account for their enduring impact on future conversations.

\subsubsection{Reward Design}
The reward function is central to the RL process, as it provides essential training signals for policy optimization. In RLFF-ESC, the reward function consists of two parts: the \textit{thinking format reward} (\(R_{\text{thk-fmt}}\)), which ensures the model generates both the intermediate reasoning process and the final supportive response, and the \textit{future-oriented reward} (\(R_{\text{fut}}\)), a neural reward model that predicts the impact of the final response on the future dialogue trajectory.

\paragraph{Thinking Format Reward.} Inspired by the recent DeepSeek-R1 model \cite{guo:2025}, which demonstrates excellent performance across reasoning tasks~\cite{xie:2025,jin:2025} through its rule-based reward design. 
We extend this approach in RLFF-ESC by introducing \(R_{\text{thk-fmt}}\), which evaluates whether the output from the LLM-based ESC model adheres to a predefined structural format. Specifically, we fine-tune the model using a prompt (see Appendix A.1) that instructs the model to enclose its reasoning process and the response within special tokens \texttt{<think>} \texttt{</think>} and \texttt{<response>} \texttt{</response>}, respectively. This structured format facilitates the model outputting its immediate reasoning process when generating a response. The thiking format reward is formally defined as follows: 

\begin{equation}
\mathcal{R}_{\text{thk-fmt}} =
\begin{cases}
1, & \text{if the output adheres to} \\
   & \text{the defined format;} \\
0, & \text{otherwise.}
\end{cases}
\end{equation}

\paragraph{Future-oriented Reward.} To predict \(R_{\text{fut}}\), we train a neural reward model to estimate the enduring emotional impact of LLMs' generated responses. This outcome-level reward implicitly guides the optimization of the LLMs’ reasoning process during reinforcement learning.
Specifically, we train the reward model using the dataset \(\mathcal{D}_r\) from the dialogue simulation. Each instance in \(\mathcal{D}_r\) consists of a conversation context \(c_{i}\), a corresponding system response \(u^{sys}_i \), and an associated reward \( r^{sys}_i\). 
To train the reward model, each instance is transformed into a training pair \( (s_i, y_i) \), where \( s_i \) represents the input sequence and \( y_i \in \{0, 1\} \) is a binary label indicating whether the system response effectively addresses the user's emotional problem or not.
Specifically, the label \( y_i \) is derived from the scalar reward \( r^{sys}_i \) using a predefined threshold \( \delta \). The label is set to one if \( r^{sys}_i \) is greater than \(\delta\); otherwise, it is set to zero.

Each input \( s_i \) is constructed using a prompt template (see Appendix A.1), incorporating the conversation context and the system response \( u^{sys}_i \). 
This input is then fed into an LLM-based classifier, which consists of a frozen LLaMA model \cite{grattafiori:2024} followed by a linear layer, to output logits \( \mathbf{z}_i \):
\begin{equation}
    \mathbf{z}_i = \text{Linear}(\text{Pool}(\text{LLaMA}(s_i))) \in \mathbb{R}^2,
\end{equation}
where a pooling layer \( \text{Pool}(\cdot) \) is applied to aggregate token-level hidden states into a single vector.
The predicted label \(\hat{y}_{i}\) is derived using a sigmoid activation function:
\begin{equation}
    \hat{y}_{i} = \sigma(\mathbf{z}_{i}),
\end{equation}
where \(\sigma(\cdot)\) is the sigmoid function.
For training, we minimize the cross-entropy loss:
\begin{equation}
    \mathcal{L} = -\frac{1}{N} \sum_{i=1}^{N} \left[ y_{i} \log(\hat{y}_{i}) + (1 - y_{i}) \log(1 - \hat{y}_{i}) \right],
\end{equation}
where \(N\) is the total number of training samples.
\(R_{\text{fut}}\) denotes the prediction made by the trained LLM classifier.

The final reward value, denoted as \( R_{\text{rlff}} \), is computed as the weighted sum of the long-term reward \(R_{\text{fut}}\) and the format reward \(R_{\text{thk-fmt}}\):
\begin{equation}
    R_{\text{rlff}} = R_{\text{fut}} + \alpha R_{\text{thk-fmt}},
\end{equation}
where \( \alpha \) is a weight controlling the importance of \( R_{\text{thk-fmt}} \) relative to \( R_{\text{fut}} \).

\subsubsection{RL Fine-tuning with GRPO}
We employ GRPO~\cite{shao:2024} to optimize the LLM model using the designed reward function \(R_{\text{rlff}}\). Compared to RL algorithms used in RLHF, such as PPO \cite{schulman:2017}, GRPO is shown to be able to simplify and stabilize the training process. This is achieved by introducing advantage normalization across groups of candidate responses, thus eliminating the need for a critic model. 

To be specific, in our RLFF-ESC framework, for each conversation context \(c_{1:t-1}\), GRPO samples a group of candidate outputs \(({o_{1}, o_{2}, \dots, o_{G}})\) from the old policy \(\pi_{\text{old}}\), which is a LLM-based ESC model. The corresponding rewards \(({r_{1}, r_{2}, \dots, r_{G}})\) are then obtained by computing the final reward of each response \(o_{i}\) using \(R_{\text{rlff}}\). To determine the relative quality of each response, the normalized advantage can be defined as follows:
\begin{equation}
    A_{i} = \frac{r_{i} - \text{mean}(r_{1}, \dots, r_{G})}{\text{std}(r_{1}, \dots, r_{G})}.
\end{equation}
The policy model \(\pi_{\theta}\) is then optimized by maximizing the following objective function:
\begin{align}
\mathcal{J} & _{\text{GRPO}}(\theta) = \mathbb{E}_{c_{1:t-1} \sim \mathcal{D}, \{o_{i}\}_{i=1}^{G} \sim \pi_{\theta}(\cdot \mid c_{1:t-1})} \small \frac{1}{G}\sum_{i=1}^{G} \nonumber \\
& \bigg[
\small \min\bigg(\frac{\pi_{\theta}(o_{i}|c_{1:t-1})}{\pi_{\text{old}}(o_{i}|c_{1:t-1})} A_{i}, \text{clip}\bigg(\frac{\pi_{\theta}(o_{i}|c_{1:t-1})}{\pi_{\text{old}}(o_{i}|c_{1:t-1})}, \nonumber \\ 
& \small 1 - \epsilon, 1 + \epsilon\bigg) A_{i} \bigg) - \beta \mathbb{D}_{\text{KL}}(\pi_{\theta} \| \pi_{\text{ref}})\bigg], 
\end{align}
where \(\epsilon\) controls the clipping range to ensure stable update, and \(\beta\) penalizes the deviation from the reference policy \(\pi_{\text{ref}}\).

\section{Experiments}

\subsection{ Experimental Setups}
\paragraph{Datasets.} 
We evaluate our method on two ESC datasets: \textbf{ESConv}~\cite{liu:2021} and \textbf{ExTES}~\cite{zheng:2023}. The ESConv dataset contains 1,300 crowd-sourced dialogues with 8 emotional support strategies, along with user problem types, emotion types, and situation descriptions.
We use the official split\footnote{\url{https://huggingface.co/datasets/thu-coai/esconv}}. 
The ExTES dataset comprises 11,177 ChatGPT-generated dialogues, verified by human annotators, covering 16 emotional support strategies. We randomly split the dataset into train/dev/test with an 8:1:1 ratio.

\paragraph{Evaluation Metrics.} 
Following prior work~\cite{deng:2024}, we adopt both automatic and human evaluation metrics. For automatic evaluation, we use two metrics: \textit{Success Rate} (SR) and \textit{Average Turn} (AT). \textit{SR} measures the proportion of dialogues in which the model successfully achieves the predefined goal within a maximum number of dialogue turns. 
\textit{AT} calculates the average number of turns required to reach the goal, reflecting the model's efficiency in task completion. For human evaluation, the quality of generated responses is assessed from five perspectives: \textit{Fluency}, \textit{Empathy}, \textit{Identification}, \textit{Suggestion}, and \textit{Overall}. 

\paragraph{Implementation Details.} 
We adopt LLaMA-3.1-8B-Instruct~\cite{grattafiori:2024} and Qwen2.5-7B-Instruct-1M~\cite{Yang:2025} as backbone models for training.
For the long-term reward model, we utilize LLaMA-3.2-1B~\cite{grattafiori:2024}. 
For evaluation, we follow the protocol of previous work~\cite{deng:2024} and use GPT-4o~\cite{achiam:2023} to role-play both the user simulator and the critic agent. 
Prompts are presented in Appendix A.2 and A.3. More details are provided in Appendix B.

\begin{table*}[ht]
\centering
\scalebox{0.9}{
\begin{tabular}{l|cc|cc|cc|cc}
\hline
\multirow{3}{*}{Method} & \multicolumn{4}{c|}{ESConv} & \multicolumn{4}{c}{ExTES} \\ \cline{2-9} 
                               & \multicolumn{2}{c|}{LLaMA-3.1} & \multicolumn{2}{c|}{Qwen2.5} & \multicolumn{2}{c|}{LLaMA-3.1} & \multicolumn{2}{c}{Qwen2.5} \\ \cline{2-9} 
                               & SR (\%) $\uparrow$ & AT $\downarrow$ & SR (\%) $\uparrow$  & AT $\downarrow$ & SR (\%) $\uparrow$ & AT $\downarrow$ & SR (\%) $\uparrow$ & AT $\downarrow$\\ \hline
Standard Prompt  & 6.15  & 7.94  & 16.9  & 7.78  & 10.3  & 7.88  & 10.3  & 7.88  \\
ESCoT            & 8.46  & 7.95  & 27.7  & 7.59  & 11.9  & 7.87  & 22.2  & 7.64  \\ 
SFT              & 6.14  & 7.82  & 8.46  & 7.78  & 13.5  & 7.87  & 11.9  & 7.86  \\ 
EmoDynamiX       & 10.8  & 7.85  & 19.2  & 7.67  & -  & -  & -  & -       \\ 
PPDP             & 20.8  & 7.81  & 26.9  & 7.55  & -  & -  & -  & -       \\
Ask-an-Expert    & 16.9  & 7.80  & 21.5  & 7.65  & 17.5  & 7.78  & 26.2  & 7.60  \\ 
GPD-Zero         & 22.9  & 7.23  & 27.9  & 7.55  & -  & -  & -  & -      \\
ICL-AIF          & \underline{23.4}  & \underline{7.18}  & \underline{28.5}  & \underline{7.43}  & \underline{28.4}                                             & \underline{7.48}  & \underline{30.3}  & \underline{7.34}  \\ \hline
\textbf{RLFF-ESC}                  & \textbf{35.5}  & \textbf{6.83}  & \textbf{41.5}  & \textbf{7.18}  & \textbf{30.2}  &  \textbf{7.30}    
                                     & \textbf{32.5 } & \textbf{7.29} \\ \hline
\end{tabular}}
\caption{Experimental results on two ESC datasets. The best results are \textbf{bolded} and the second-best results are \underline{underlined}. SR denotes the Success Rate, and AT represents the Average Turns to reach the goal.}
\label{table:exp_res}
\end{table*}

\begin{table*}[h]
    \centering
    \scalebox{0.9}{
    \begin{tabular}{lcccccccccc}
    \toprule
    RLFF-ESC & \multicolumn{2}{c}{Fluency} &  \multicolumn{2}{c}{Empathy} &  \multicolumn{2}{c}{Suggestion} & \multicolumn{2}{c}{Identification} & \multicolumn{2}{c}{Overall} \\
    \cmidrule(lr){2-3} 
    \cmidrule(lr){4-5} 
    \cmidrule(lr){6-7} 
    \cmidrule(lr){8-9} 
    \cmidrule(lr){10-11} 
    \multicolumn{1}{c}{vs.} & Win & Lose & Win & Lose & Win & Lose & Win & Lose & Win & Lose \\
    \midrule
    Standard Prompt & 96\% & 4\%  & 92\% & 8\%  & 90\% & 10\% & 92\%  & 8\%   & 92\% & 8\% \\
    ESCoT           & 92\% & 8\%  & 88\% & 12\% & 84\% & 16\% & 88\%  & 12\%  & 88\% & 12\% \\
    SFT	            & 84\% & 16\% & 78\% & 22\% & 78\% & 22\% & 82\%  & 18\%  & 82\% & 18\% \\
    EmoDynamiX      & 98\% & 2\%  & 96\% & 4\%  & 96\% & 4\%  & 98\%  &2\%    & 98\% & 2\%  \\
    PPDP            & 94\% & 6\%  & 94\% & 6\%  & 98\% & 2\%  & 98\%  & 2\%   & 96\% & 4\% \\
    Ask-an-Expert   & 74\% & 26\% & 82\% & 18\% & 78\% & 22\% & 78\%  & 22\%  & 76\% & 24\% \\
    GPD-Zero        & 88\% & 12\% & 86\% & 14\% & 86\% & 14\%  & 84\% & 16\%  & 82\% & 18\% \\
    ICL-AIF         & 66\% & 34\% & 78\% & 22\% & 76\% & 24\%  & 76\% & 24\%  & 74\% & 26\% \\
    \bottomrule
    \end{tabular}}
    \caption{Human evaluation results comparing RLFF-ESC with the baselines across five metrics.}
    \label{tab:human_eval}
\end{table*}

\paragraph{Baselines.}
We compare our method with a range of baselines:
(1) Prompt-based response generation methods that directly prompt LLMs to generate emotionally supportive responses. This category includes \textbf{Standard Prompt}~\cite{deng:2024} and \textbf{ESCoT}~\cite{zhang_escot:2024}.
(2) Supervised fine-tuning (\textbf{SFT})~\cite{zheng:2023} that fine-tunes LLMs on ESC datasets to enhance the generation of supportive responses.
(3) Policy planning with external modules, which integrates LLMs with external policy models to predict dialogue strategies and generate responses. This category includes \textbf{PPDP}~\cite{deng:2024}, \textbf{EmoDynamiX}~\cite{wan:2025}, and \textbf{Ask-an-Expert}~\cite{zhang:2023}.
(4) Prompt-based iterative planning and feedback, which utilizes multi-turn simulations to provide strategic foresight or planning signals that guide the response generation process. This category includes \textbf{ICL-AIF}~\cite{fu:2023} (dialogue-level feedback via simulation) and \textbf{GPD-Zero}~\cite{yu:2023} (planning via MCTS with LLM-based components). 
More details can be found in the Appendix C.

\subsection{Experimental Results}
\subsubsection{Overall Performance}
\paragraph{Comparison with baselines.}

\begin{figure}[h]
    \centering
    \includegraphics[width=0.47\textwidth]{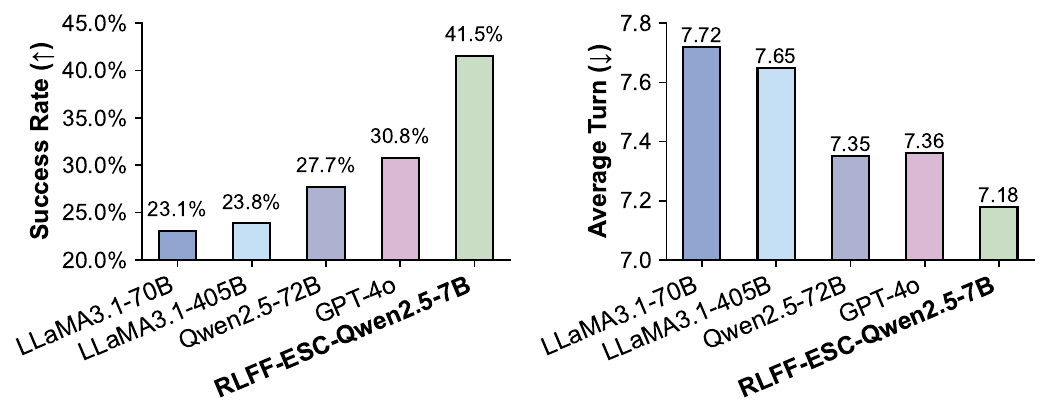}
    \caption{Performance comparison of RLFF-ESC and larger-scale LLMs: the left shows Success Rate, and the right shows the Average Turn performance.}
    \label{fig:comparison_llms}
\end{figure}

It shows that our approach, RLFF-ESC, consistently outperforms all baselines across both datasets in terms of \textit{success rate} (SR) and \textit{average number of turns} (AT). These results demonstrate that optimizing LLMs by considering the long-term impact of each individual response on future dialogue trajectories can improve the performance of goal completion, thus facilitating higher-quality emotional support.  Furthermore, our method exhibits superior performance in emotional support across different LLM backbones, indicating its generalizability and adaptability. Specifically, when being adapted to the Qwen2.5-7B-Instruct-1M model, our approach achieves better overall performance compared to its integration with the LLaMA3.1-8B-Instruct model.

\begin{figure*}[ht]
    \centering
    \includegraphics[width=\textwidth]{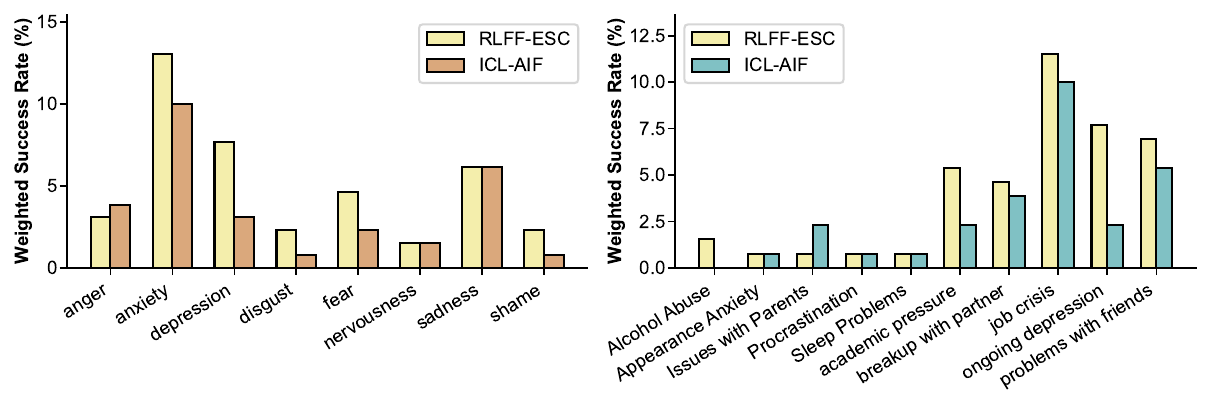}
    \caption{Performance comparison between RLFF-ESC and the second-best method, ICL-AIF, on the ESConv dataset across different user emotion types (left) and problem types (right). RLFF-ESC more effectively alleviates user distress in most cases.}
    \label{fig:emo_problem}
\end{figure*}

\paragraph{Comparison with Larger-Scale LLMs.}
We compare the performance of RLFF-ESC with several representative large-scale LLMs on the ESConv dataset, including GPT-4o, LLaMA-3.1-70B-Instruct, LLaMA-3.1-405B-Instruct and Qwen2.5-72B-Instruct, in a zero-shot setting using the same prompt for the sake of consistency. Figure~\ref{fig:comparison_llms} summarizes the results. 
Our method, based on a compact 7B model, achieves a success rate of $41.5\%$, outperforming LLaMA-3.1-405B-Instruct ($23.9\%$), Qwen2.5-72B-Instruct ($27.7\%$), and GPT-4o ($30.8\%$). 
Despite being based on a compact 7B model, RLFF-ESC achieves a success rate of $41.5\%$, surpassing LLaMA-3.1-405B-Instruct ($23.9\%$), Qwen2.5-72B-Instruct ($27.7\%$), and GPT-4o ($30.8\%$).
RLFF-ESC also generates emotional support conversations with fewer average turns.

\paragraph{Human Evaluation.}
In line with previous work~\cite{peng:2022, zhao_transesc:2023, deng:2024}, we conduct a human evaluation on 100 dialogues randomly sampled from the ESConv dataset. 
Three annotators with backgrounds in psychology were instructed to compare the responses generated by our method and the baselines across five dimensions (see detailed instructions in Appendix D.1).
As shown in Table~\ref{tab:human_eval}, RLFF-ESC consistently outperforms the baselines on all human evaluation metrics. This indicates that our approach can generate more effective emotionally supportive responses, especially in terms of empathy, fluency, problem identification, and offering suggestions to help users work through their challenges.
A representative case study of our method is shown in Table~\ref{tab:case_study} (see more examples in Appendix D.2).
To further validate the reliability of the LLM-based critic agent, we conduct agreement analysis with human preferences (see Appendix D.3). 

\subsubsection{Ablation Study}

\begin{table}[h]
\centering
\scalebox{1}{
\begin{tabular}{lcc}
\toprule
Qwen2.5-7B & SR (\%) $\uparrow$ & AT $\downarrow$ \\
\midrule
+ vanilla   & 26.5  & 7.58  \\  
+ DPO       & 31.5  & 7.26   \\ 
+ GRPO\_\textit{Ranking}   & 26.9   & 7.56  \\ 
+ GRPO\_\textit{Random}     & 18.5   & 7.66   \\ 
\midrule
+ GRPO\_\textit{Classification}   & 41.5   & 7.18  \\
\bottomrule
\end{tabular}}
\caption{Experimental results of the ablation study on the ESConv dataset. \textit{Ranking}, \textit{Random}, and \textit{Classification} indicate using ranking, random, and classification reward models, respectively, for the future-oriented reward prediction in GRPO.}
\label{tab:ablation}
\end{table}

To evaluate the key components of RLFF-ESC, we conducted an ablation study on the ESConv dataset with Qwen2.5-7B-Instruct-1M as the backbone. The results are shown in Table~\ref{tab:ablation}. 
To evaluate the effectiveness of GRPO, we remove it and apply the same vanilla prompt as RLFF-ESC in a zero-shot setting, resulting in a significant performance decline.
We then replace GRPO with DPO, using data constructed during the multi-agent dialogue simulation. More specifically, for a group of (response-reward) pairs, the highest reward response is ``chosen," and the lowest is ``rejected." DPO improves performance over the vanilla setting, proving the usefulness of multi-agent dialogue simulation, but it still underperforms compared to GRPO. 
Recognizing the importance of the reward model for GRPO, we further tested different designs. First, using an untrained LLM for future-oriented reward prediction, i.e., GRPO\_\textit{Random}, leads to a dramatic performance drop, implying the need for a suitable reward function in RL. Next, we train a LLM-ranking model with RRHF~\cite{Yuan:2023} on the constructed dataset. This approach yields performance similar to that of the vanilla method, suggesting that a reward model based on classification might provide straightforward guidance than the more complex ranking method.

\subsubsection{Further Analysis}
In addition, Figure~\ref{fig:emo_problem} shows the performance of RLFF-ESC with the second-best model ICL-AIF across different user emotion types and problem types. We report weighted success rates, reflecting each category’s proportion in the dataset. Results show that RLFF-ESC outperforms ICL-AIF in most emotion types and problem types, particularly in cases involving \textit{anxiety}, \textit{depression}, \textit{ongoing depression}, and \textit{academic pressure}. These findings highlight the effectiveness of our model in various scenarios. For more experimental analysis, please refer to the Appendix E.

\begin{table}[hb]
    \centering
    \scalebox{0.85}{
    \begin{tabular}{p{8.5cm}}
    \toprule
    \textbf{Conversation Context (Input): }    \\
    \textbf{ASSISTANT:} \textit{``Hi how are you today?''}   \\
    \textbf{USER:} \textit{``OK a little upset''}   \\
    \textbf{ASSISTANT:} \textit{``I'm sorry. Can you tell me what happened?''}  \\
    \textbf{USER:} \textit{``I thought I was in a good relationship and just like that he calls it quits. We went on a date and he just decided to break up with me.''}  \\
    \midrule
    \textbf{RLFF-ESC Response:} \texttt{<think>} The patient is experiencing sudden emotional distress from an unexpected breakup. They may feel rejected, confused, and hurt. It's important to validate their feelings, offer empathy, and gently encourage them to share more about their experience and emotions. \texttt{</think>} \texttt{<response>} That sounds really painful and confusing, especially when it felt like things were going well. It’s completely understandable to feel upset after something so sudden. Would you like to talk more about what happened on the date or how you’re feeling right now? \texttt{</response>}\\
    \bottomrule
    \end{tabular}}
    \caption{ A Case study of RLFF-ESC.}
    \label{tab:case_study}
\end{table}

\section{Conclusion}
In this paper, we propose RLFF-ESC, a novel end-to-end framework that improves LLMs through reinforcement learning for open-ended emotional support generation with enduring emotional impact.
We employ an effective reward mechanism, comprising thinking format and future-oriented rewards, to guide LLM optimization via GRPO.
Extensive experiments indicate that RLFF-ESC improves both task completion rates and response quality in a variety of emotional support scenarios. Moreover, evaluations with different LLM backbones highlight its adaptability. 
Further analysis reveals that the simple reward formulation (i.e., the classification-based) can be more effective than complex ones (e.g., the ranking-based) for future-oriented reward modeling.

\section*{Limitations}
While our method outperforms existing baselines on ESC tasks, there remains a gap between its current performance and the requirements for practical applications. For automatic evaluation, we utilize LLMs, For automatic evaluation, we utilize LLMs because they have demonstrated strong performance in terms of user simulation. However, potential evaluation biases within LLMs may affect the results' reliability. Although we conducted human evaluations, our study did not 
assess changes in end users’ emotional intensity. In future work, we plan to conduct comprehensive user studies to evaluate the practical effectiveness of our proposed method in real-life situations.

\section*{Ethical Considerations}
It is important to clarify that the term \textit{emotional support conversation} in this paper is intended to simulate support through social interactions (e.g., from peers or friends) rather than professional counseling or psychological treatment. Although terms such as \textit{``therapist''} and \textit{``patient''} appear in the prompts used, they are solely for illustrative purposes and do not indicate the provision of clinical or therapeutic services. This study does not involve any form of professional counseling or mental health intervention. Additionally, all datasets utilized in this research were obtained from publicly available sources. These datasets do not contain any personally identifiable or sensitive information about users.



\bibliography{custom}

\appendix
\section{Prompting Details} 
This section outlines the details of the prompts used in our implementation.

\subsection{RL Tuning Prompt} 
\label{subsec:appendix_rl_prompt}
We begin by describing the prompt used for reinforcement learning with GRPO. The prompt template for training a reward model for future-oriented reward estimation is represented in Table~\ref{tab:prompt_template}. Moreover, to guide the model in generating reasoning processes and responses in a predefined format, we utilize a system prompt, as illustrated in Table \ref{tab:rl_task_prompt}. 

\begin{table}[h]
\centering
\begin{tabular}{p{0.95\linewidth}}
\toprule
\textbf{Instruction:} You are a helpful assistant tasked with evaluating whether a patient's emotional problem has been effectively addressed following a conversation with a therapist. \\

\textbf{Input Format:} \\
Conversation: \{conversation context\} \\
Therapist's response: \{system response\} \\
\bottomrule
\end{tabular}
\caption{Prompt template for training a reward model for future-oriented reward estimation.}
\label{tab:prompt_template}
\end{table}

\begin{table*}[h]
    \centering
    \renewcommand{\arraystretch}{1.3}
    \begin{tabular}{p{13.5cm}}
    \toprule
    Now enter the role-playing mode. In the following conversation, you will play as a therapist in a counselling conversation with a patient. Your goal is to help the patient reduce their emotional distress and support them working through their challenges. You first think about the reasoning process in the mind and then provide the patient with the response. The reasoning process and response are enclosed within \texttt{<think> </think> }and \texttt{<response> </response>} tags, respectively, i.e., \texttt{<think>} reasoning process here \texttt{</think> <response>} supportive response here \texttt{</response>}. \\
    \bottomrule
    \end{tabular}
    \caption{Prompt template for RL tuning.}
    \label{tab:rl_task_prompt}
\end{table*}

\subsection{User Simulator Prompt}
\label{subsec:appendix_u_prompt}
We prompt the LLMs to role-play as users using the prompt proposed by. This prompt instructs the LLMs to simulate users, specifying the emotion type (which is set to None in the ExTES dataset) and the problem type. The prompt is detailed in Table \ref{tab:user_simulator_prompt}.

\begin{table*}[h]
    \centering
    \begin{tabular}{lp{11cm}}
    \toprule
    System   &  Now enter the role-playing mode. In the following conversation, you will play as a patient in a counselling conversation with a therapist. \\
    \midrule
    User    &  You are the patient who is looking for the help from the therapist, because you have the emotional issue about \textbf{[emotion\_type]} regarding \textbf{[problem\_type]}. Please reply with only one short and succinct sentence. Now tell me your issue.\\
    \midrule
    Assistant & \textbf{[situation]} \\
    \bottomrule
    \end{tabular}
    \caption{Prompt for user simulator.}
    \label{tab:user_simulator_prompt}
\end{table*}

\begin{table}[h]
    \centering
    \scalebox{0.85}{
    \begin{tabular}{lll}
    \toprule
    Training Phase   &  Hyper-parameter & Value\\
    \midrule
     \multirow{5}{*}{Reward Model}  & Batch Size & 1\\
     & Training Epochs & 2 \\
     & Learning Rate & 1e-4\\
     & Max Sequence Length & 2048 \\
     & Gradient Accumulation Steps & 8 \\
    \midrule
    \multirow{9}{*}{RL} & Batch Size & 4\\
    & Training Epochs & 2\\
    & Learning Rate & 1e-6\\
    & Lora Rank & 8 \\
    & Lora alpha & 32 \\
    & Max Dialogue Turn & 8 \\
    & Number of Generations & 4 \\
    & Temperature & 1.1 \\
    & Top p & 1.0 \\
    & Top k & 80 \\
    \bottomrule
    \end{tabular}}
    \caption{Hyper-parameter settings.}
    \label{tab:hyperparam}
\end{table}

\subsection{Critic Agent Prompt}
\label{subsec:appendix_critic_prompt}
The prompts for the critic agent are designed to evaluate the degree of achievement of the goal within the dialogue. To facilitate a more nuanced and fine-grained assessment, we extend the previous four-level reward scale to seven levels. This fine granularity enables the critic model to better distinguish subtle differences in goal achievement, as detailed in Table~\ref{tab:critic_prompt}.

\begin{table*}[h]
    \centering
    \begin{tabular}{lp{14.5cm}}
    \toprule
    System   &  Given a conversation between a therapist and a patient, please assess whether the patient' emotional issue has been solved after the conversation. \\
    \midrule
    User     & The following is a conversation about \textbf{[emotion\_type]} regarding \textbf{[problem\_type]}: \textbf{[conversation]} \\
    & \textbf{Question}: Has the patient's issue been solved? \\
    & \textbf{Requirement}: You can only reply with one of the following seven descriptive levels without explanation: \\
    & \textit{Same:} \textit{The patient's feelings remain unchanged.} \\
    & \textit{Slightly Better:} \textit{The patient feels a slight, barely noticeable improvement.} \\
    & \textit{Moderately Better:} \textit{The patient feels somewhat better, with a small but noticeable improvement.} \\
    & \textit{Significantly Better:} \textit{The patient feels significantly better, indicating a clear improvement.}\\
    & \textit{Slightly Worse:} \textit{The patient feels a slight increase in tension, barely noticeable decline.} \\
    & \textit{Moderately Worse:} \textit{The patient feels somewhat more stressed or worried, with a small but noticeable decline.} \\
    & \textit{Significantly Worse:} \textit{The patient feels significantly more distressed or upset, indicating a clear decline.} \\
    \bottomrule
    \end{tabular}
    \caption{Prompt for the critic agent.}
    \label{tab:critic_prompt}
\end{table*}

\section{Implementation Details}
\label{sec:appendix_train_detail}
\paragraph{Training Details.} 
To construct the data for training the future-oriented reward model of RLFF-ESC, the Qwen-2.5-7B-Instruct-1M model is prompted to role-play as the ESC system \(M_{sys}\) in the multi-agent dialogue simulation process. For each conversation context, a set of \(n=4\) responses is sampled by the ESC system by setting: \(\tau = 1.1\), \(top\_k = 80\), and \(top\_p = 1.0\). To reduce training costs, we replace GPT-4o-2024-11-20 (used during evaluation) with Qwen-2.5-72B-Instruct to role-play both the user simulator \(U\) and the critic \(\mathcal{M}_{\text{crt}}\) during dialogue simulations. 

We fully fine-tune LLaMA-3.2-1B as an LLM-based classifier to predict the future-oriented reward using the constructed data. We use 
LLaMA-3.1-8B-Instruct and Qwen-2.5-7B-Instruct-1M as the backbone models of our method and optimize them using the GRPO reinforcement learning algorithm with LoRA. The reward weights of \(R_{\text{fut}}\) and \(R_{\text{thk-fmt}}\) are \(1\) and \(0.5\), respectively. All experiments are implemented in PyTorch and conducted on 4 NVIDIA A100 GPUs with the DeepSpeed library using ZeRo-3 optimization. Detailed hyperparameter settings are provided in Table~\ref{tab:hyperparam}.

\paragraph{Evaluation Details.} 
For evaluation, we follow the protocol of previous work and use GPT-4o to role-play both the user simulator and the critic agent. The critic agent assesses whether the user's problem is resolved or not. We set temperature \(\tau = 1.1\) and sample feedback \(l = 10\) times. Feedback levels include: \textit{``significantly worse,'' ``moderately worse,'' ``slightly worse,'' ``same,'' ``slightly better,'' ``moderately better,''} and \textit{``significantly better,''} mapped to \(-1.0\), \(-0.5\), \(-0.25\), \(0\), \(0.25\), \(0.5\), and \(1.0\), respectively. We aggregate the sampled feedback to compute a scalar reward. The dialogue goal is considered complete if the reward exceeds \(0.5\).

\section{Baselines}
\label{sec:appendix_baselines}
We reproduce all the baseline methods using their official implementations and configurations. For consistency, all methods are adapted to two base LLMs: LLaMA-3.1-8B-Instruct and Qwen-2.5-7B-Instruct-1M.

\begin{itemize}
    \item \textbf{Standard Prompt:} The LLM is prompted to generate emotionally supportive responses directly, without intermediate reasoning steps.

    \item \textbf{ESCoT:} A chain-of-thought prompting method that guides the LLM to reason through the user's emotional state, the triggering event, emotion appraisal, and an appropriate supportive strategy before composing a response.

    \item \textbf{SFT:} As many recent approaches improve emotional support capabilities through SFT, we fine-tune both base LLMs on existing ESC datasets.

    \item \textbf{PPDP:} A policy planning method using a RoBERTa-based model trained to predict the next dialogue strategy. The model is optimized via reinforcement learning using AI-generated feedback.

    \item \textbf{EmoDynamiX:} A RoBERTa-based policy model that incorporates a heterogeneous graph to model complex discourse dynamics between user emotions and system strategies.

    \item \textbf{Ask-an-Expert:} The LLM is instructed to act as a strategy expert, reasoning about the most appropriate response strategy given the conversation context.

    \item \textbf{ICL-AIF:} Two LLMs engage in a role-play emotional support conversation (e.g., seeker vs. supporter), while a third LLM provides feedback to improve each agent's planning and strategy. This setup encourages forward-looking reasoning and iterative strategy refinement.

    \item \textbf{GPD-Zero:} The LLM is used as components like the prior policy in a Monte Carlo Tree Search framework for planning goal-oriented dialogue strategies. In our implementation, we replace ChatGPT with our two base models to ensure consistent evaluation.
\end{itemize}

\begin{figure*}[h]
    \centering
    \begin{subfigure}[b]{0.45\textwidth}
        \includegraphics[width=\textwidth]{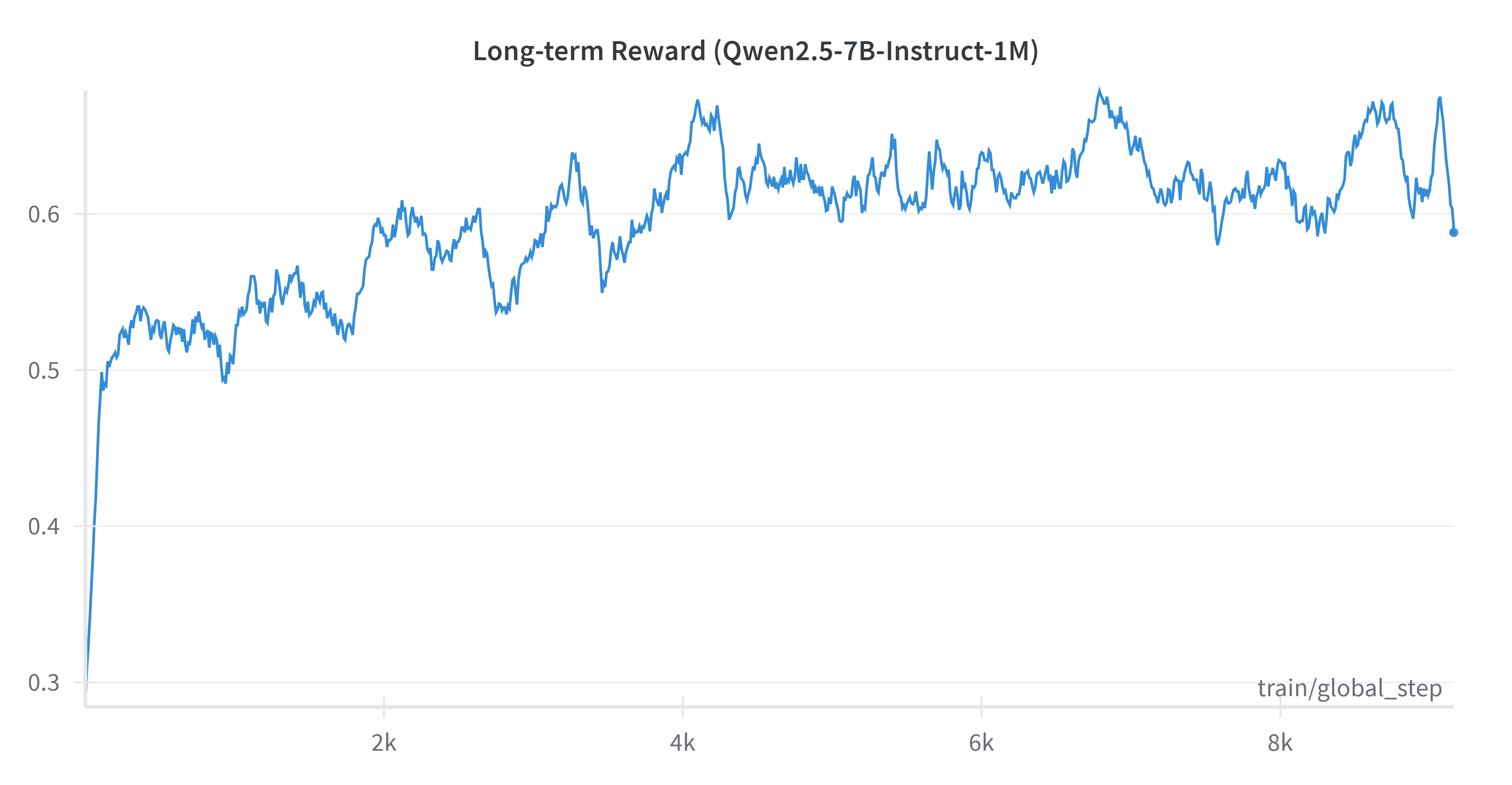}
        \caption{}
        \label{fig:sub1}
    \end{subfigure}
    \hfill
    \begin{subfigure}[b]{0.45\textwidth}
        \includegraphics[width=\textwidth]{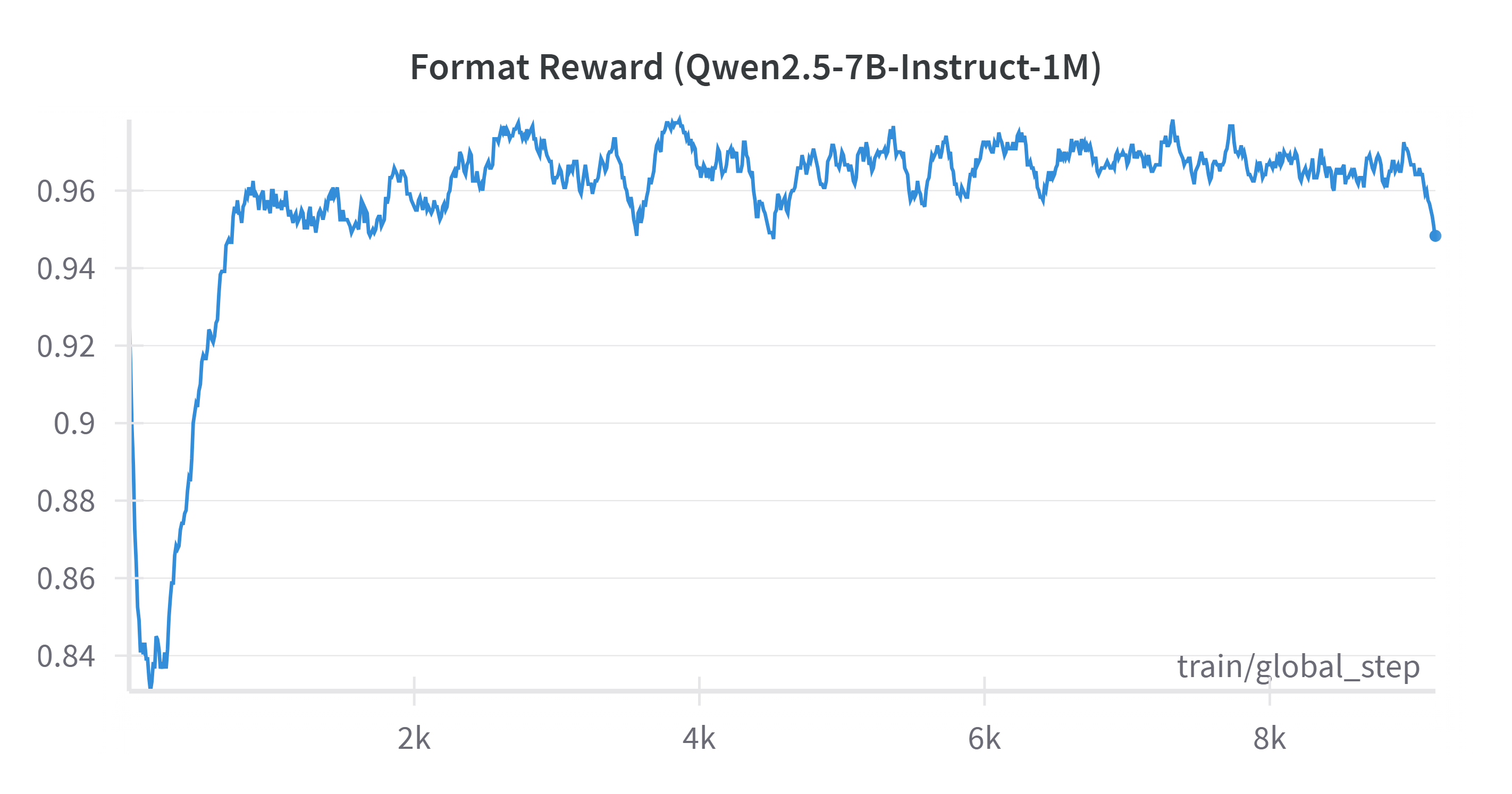}
        \caption{}
        \label{fig:sub2}
    \end{subfigure}
    
    \vskip\baselineskip

    \begin{subfigure}[b]{0.45\textwidth}
        \includegraphics[width=\textwidth]{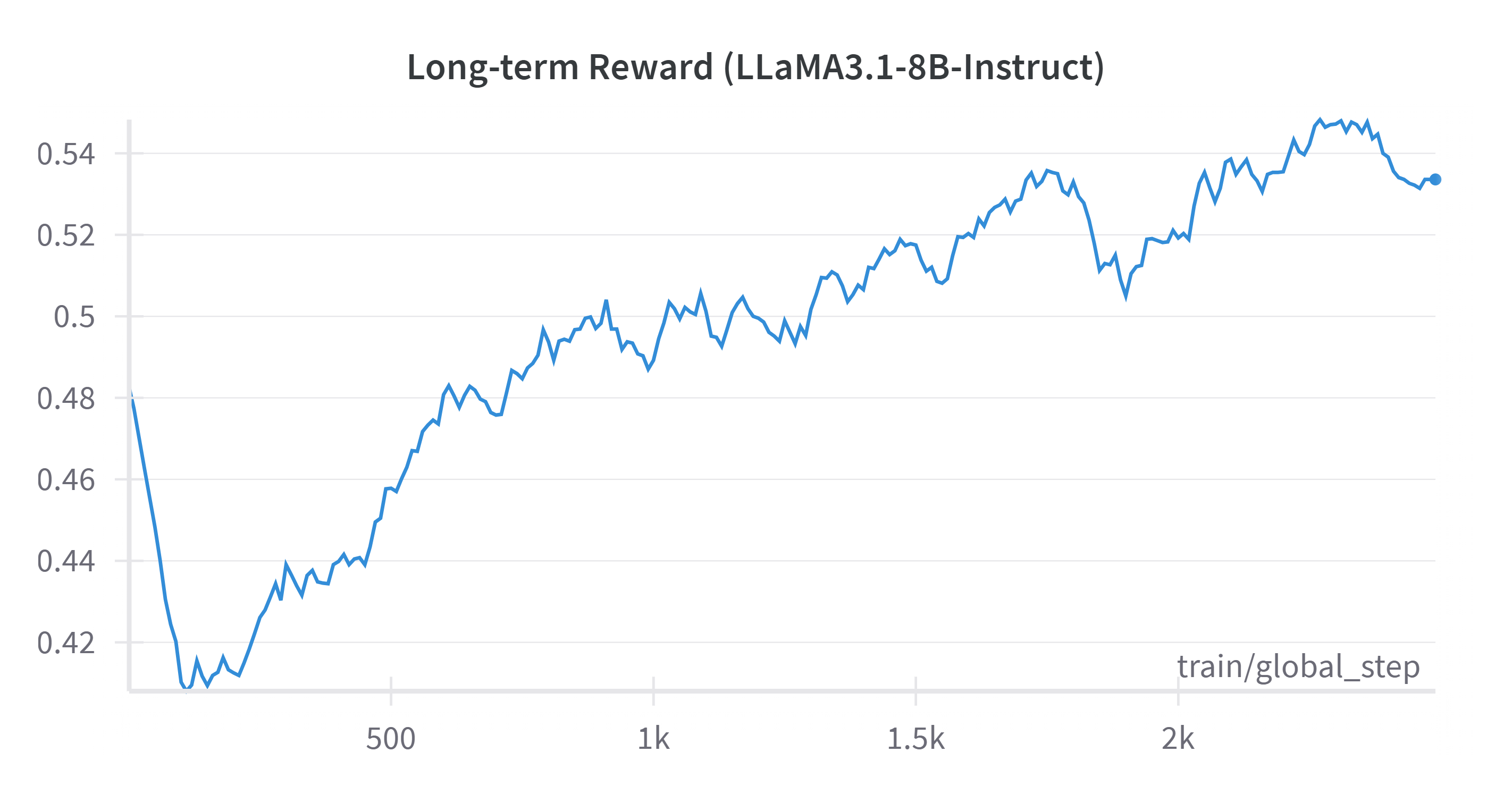}
        \caption{}
        \label{fig:sub3}
    \end{subfigure}
    \hfill
    \begin{subfigure}[b]{0.45\textwidth}
        \includegraphics[width=\textwidth]{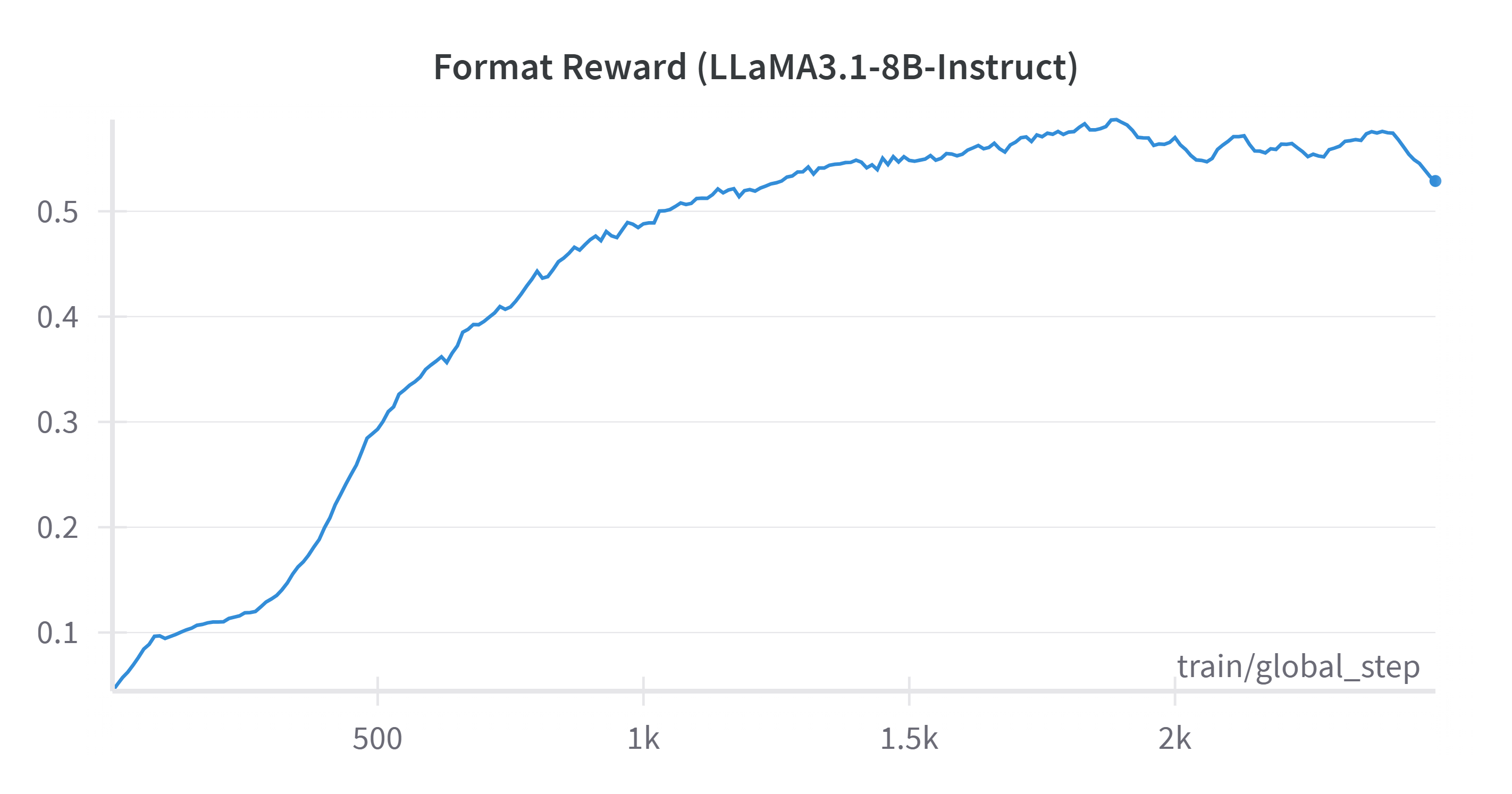}
        \caption{}
    \end{subfigure}
    
    \caption{Training Rewards. (a) Future-oriented reward during training for Qwen2.5-7B-Instruct-1M. (b) Thinking Format reward during training for Qwen2.5-7B-Instruct-1M. (c) Future-oriented reward during training for LLaMA3.1-8B-Instruct. (d) Thinking Format reward during training for LLaMA3.1-8B-Instruct.}
    \label{fig:training_reward}
\end{figure*}

\begin{figure*}[h]
    \centering
    \includegraphics[width=\textwidth]{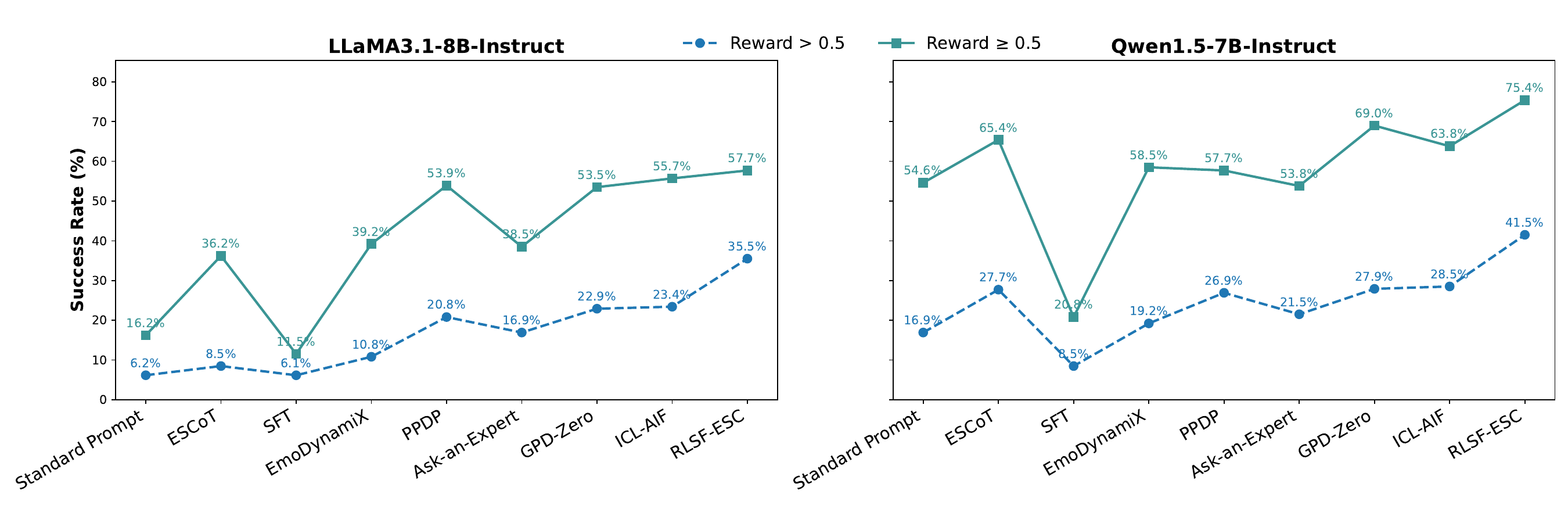}
    \caption{Model performance of RLFF-ESC and baselines on different evaluation criteria on the ESConv dataset.}
    \label{fig:evaluation_thresholds}
\end{figure*}


\section{Human Evaluation}
\label{sec:appendix_human_eval}
\subsection{Response Evaluation Instructions} 
We recruited three annotators with a background in psychology. They were tasked with comparing the responses generated by our method and the baselines based on four primary perspectives, as well as providing an overall assessment, as detailed below:
\begin{itemize}

\item \textbf{Fluency:} Which model produces the response that is more natural, well-structured, and logically coherent, with correct grammar and smooth sentence flow?

\item \textbf{Empathy:} Which model responds with greater emotional understanding, showing warmth, compassion, or concern that aligns with the user’s feelings?

\item \textbf{Identification:} Which model explores the user’s situation more effectively to identify the problem?

\item \textbf{Suggestion:} Which model offers more relevant, practical, and emotionally sensitive suggestions that could help the user cope or take action?

\item \textbf{Overall:} Overall, which model provides better emotional support, considering empathy, understanding, helpfulness, and communication quality?
\end{itemize}

\subsection{Case Study}
In Table~\ref{tab:case_study_1} and Table~\ref{tab:case_study_2}, we present two case studies comparing the responses generated by our RLFF-ESC model and the ICL-AIF method, focusing on addressing emotional problems related to \textit{ongoing depression} and \textit{alcohol abuse}, respectively. Compared to the baseline, our model demonstrates a greater capacity for empathy and emotional resonance, more accurately identifying the user's specific concerns and providing supportive, contextually appropriate suggestions.
Furthermore, in Table~\ref{tab:example_1} and Table~\ref{tab:eample_2}, we give example dialogues generated by RLFF-ESC on two datasets involving interactions with a user simulator: one centered on \textit{sleep problems} and the other on \textit{coping with the illness of a family member}. These examples further illustrate our model's ability to engage in emotionally intelligent conversations across different scenarios.

\subsection{Agreement Analysis}
To evaluate the reliability of the LLM-based critic, we conducted an alignment analysis with human preferences. We randomly sampled 100 dialogues from the ESConv dataset. Three additional human annotators, each with a background in psychology, and the LLM-based critic independently assessed whether the patient's emotional issue had been resolved, using consistent evaluation criteria. Our analysis revealed strong agreement among the human annotators (\textit{Krippendorff’s} $\alpha = 0.798$; Spearman’s $\rho = 0.794$), as well as a high level of alignment between the LLM-based critic and the human majority vote (\textit{Krippendorff’s} $\alpha = 0.786$; Spearman’s $\rho = 0.796$, $p < 0.0001$). These results suggest that the critic agent provides feedback that is closely aligned with human evaluators.

\section{Further Analyses}
\label{sec:further_analysis}
\paragraph{Training Rewards.}
Figure~\ref{fig:training_reward} presents the training rewards of two base LLMs during the reinforcement learning process of RLFF-ESC. For Qwen2.5-7B-Instruct-1M, the future-oriented reward increases sharply to 0.5 within the first 100 training steps and then continues to rise more gradually. The format reward initially decreases during the first 200 steps, followed by a sharp increase until approximately 1,000 steps, after which it continues to increase steadily. In contrast, for LLaMA3.1-8B-Instruct, the future-oriented reward declines during the first 200 training steps and subsequently increases gradually. The format reward shows a consistent upward trend throughout the training process.

\paragraph{Effect of the Evaluation Threshold.}
Figure~\ref{fig:evaluation_thresholds} illustrates the success rates of RLFF-ESC and baseline models under two different evaluation thresholds on the ESConv dataset. Specifically, we compare two definitions of success: reward $> 0.5$ (used in our experiments) and reward $\geq 0.5$. The results show that using the more lenient threshold of reward $\geq 0.5$ leads to a significant increase in the reported success rate. This indicates that a substantial number of interactions result in a reward exactly equal to 0.5, suggesting users may feel moderately better in those cases.
This observation highlights the sensitivity of evaluation metrics to threshold settings. Despite the change in absolute success rates, the relative ranking of RLFF-ESC and the baseline models remains approximately consistent across both criteria. We adopt the stricter threshold of reward $> 0.5$ in our main experiments to ensure that success reflects a clearer and more substantial improvement in the user's emotional state.

\begin{table*}[h]
    \centering
    \begin{tabular}{p{14.5cm}}
    \toprule
    \textbf{Problem Type:} Ongoing Depression   \\
    \midrule
    \textbf{Conversation Context:}    \\
    \textbf{ASSISTANT:} \textit{``Hello, what can I help you with this evening?''}   \\
    \textbf{USER:} \textit{``I am feeling very depressed lately. Like a constant pressure that I don't know exactly what it is.''}   \\
    \textbf{ASSISTANT:} \textit{``Depression is extremely rough to deal with, I'm very sorry to hear you're going through that.''}  \\
    \textbf{USER:} \textit{``I think the days getting shorter makes it worse.''}  \\
    \midrule
    \textbf{ICL-AIF:} It sounds like the change in seasons is affecting your mood significantly. Many people find that shorter days and less sunlight can contribute to feelings of depression. How do you think these changes are specifically impacting your daily life or your thoughts? \\
    \midrule
    \textbf{RLFF-ESC:} I hear you—depression can feel incredibly heavy, and it makes sense that shorter days might be amplifying those feelings. Many people find that changes in light and the seasons affect their energy levels and mood. Would you like to talk more about what this pressure feels like for you, or we could explore how the change in seasons might be playing a role? \\
    \bottomrule
    \end{tabular}
    \caption{Case study of the generated supportive responses by our RLFF-ESC and the ICL-AIF method.}
    \label{tab:case_study_1}
\end{table*}

\begin{table*}[h]
    \centering
    \begin{tabular}{p{14.5cm}}
    \toprule
    \textbf{Problem Type:} Alcohol Abuse   \\
    \midrule
    \textbf{Conversation Context:}    \\
    \textbf{USER:} \textit{Hello, how are you?} \\
    \textbf{ASSISTANT:} \textit{``Hi there, how are you feeling?''} \\
    \textbf{USER:} \textit{``I'm alright, honestly I'm feeling rather ashamed. The reason for that and the problem that I'm facing is that I've had a drinking problem for quite a few years now, and I'm only 24 years old. Right now in my life I'm ready to seek help, but I'm having a lot of anxiety about it because I'm very worried about how others will judge me. I'm worried about how the counselors will see me, how others at a facility might view me, and any one else that I will have to share my story with when getting treatment and how they might view me. I'm ashamed of myself, my life, and the choices that I've made that have gotten me here and I guess my problem is that I just don't know how to get over these extreme anxieties and worries so that I can just finally make the step that I need so desperately and get help.''} \\
    \midrule
    \textbf{ICL-AIF:} Thank you for sharing this with me. It takes a lot of strength to confront such difficult feelings and talk about your experiences. It's completely understandable to feel ashamed and worried about judgment; many people who struggle with addiction feel the same way. It's important to remember that seeking help is a brave step towards healing and recovery.  \\
    \midrule
    \textbf{RLFF-ESC:} Thank you for sharing this with me. First, let me say that it took a lot of courage for you to open up about what you're going through, especially surrounding something as sensitive as struggling with a drinking problem. The fact that you're already considering seeking help is a huge step forward, and it's an admirable act of strength and self-awareness. \\
    \bottomrule
    \end{tabular}
    \caption{Case study of the generated supportive responses by our RLFF-ESC and the ICL-AIF method.}
    \label{tab:case_study_2}
\end{table*}

\begin{table*}[h]
    \centering
    \begin{tabular}{p{14.5cm}}
    \toprule
    \textbf{Emotion Type:} Sadness \\
    \textbf{Problem Type:} Sleep Problems \\
    \textbf{Situation:} I've had trouble sleeping the last few nights, I feel really low self-esteem. \\
    \midrule
    \midrule
    \textbf{ASSISTANT:} \textit{``I'm sorry to hear you're having such a tough time. Lack of sleep and low self-esteem can be interconnected. Could you tell me more about what's been going on in your life recently that might be contributing to these feelings? ''} \\

    \textbf{USER:}  \textit{``Lately, I've been overwhelmed with work stress and personal expectations, and it's making it harder to wind down at night.''} \\
    
    \textbf{ASSISTANT:} \textit{``It's understandable to feel overwhelmed when work and personal expectations pile up. This kind of stress can definitely make it difficult to relax and get a good night's sleep. Have you noticed any particular moments or tasks that are especially challenging for you at work? Also, how do you usually cope with stress? ''} \\
    
    \textbf{USER:} \textit{``I struggle most with tight deadlines and the fear of disappointing others, and I usually just push through without addressing how I feel.''} \\
    
    \textbf{ASSISTANT:} \textit{``It sounds like you're carrying a lot of pressure on yourself, especially around meeting deadlines and avoiding disappointment. One way to start addressing this might be by setting aside some time each day to reflect on how you're feeling. Even if it's just a few minutes, acknowledging your emotions can help you process them more effectively. What do you think about trying this approach?''}  \\

    \textbf{USER:}  \textit{``That sounds like a good idea; I think setting time for reflection and smaller steps could help me feel less trapped in the stress.''} \\
    
    \textbf{ASSISTANT:} \textit{``I'm glad you find those suggestions helpful. Setting aside time for reflection and breaking tasks into smaller steps can indeed make a significant difference. Remember, it's okay to take things one step at a time. If you notice that certain strategies work better than others, don't hesitate to adjust your approach. How do you feel about starting to incorporate these changes into your routine? ''} \\
    
    \textbf{USER:} \textit{``I feel hopeful about trying these changes and curious to see how they might help me feel more in control.''}  \\
    
    \textbf{ASSISTANT:} \textit{``I'm really glad to hear you feel hopeful and curious about trying these changes. It's natural to face some challenges along the way, but remember that each small step you take is a step towards greater control over your well-being. Let's keep an open dialogue about how things are going, so we can support you in finding the best path forward. How does that sound? ''} \\
    
    \textbf{USER:} \textit{``That sounds reassuring, and I appreciate having a space to openly share and adjust as I go.''} \\
    \midrule
    \textbf{Reward: }0.8  $\rightarrow$  Goal completed !!\\

    \bottomrule
    \end{tabular}
    \caption{An example conversation with our RLFF-ESC on the ESConv dataset.}
    \label{tab:example_1}
\end{table*}

\begin{table*}[h]
    \centering
    \begin{tabular}{p{14.5cm}}
    \toprule
    \textbf{Problem Type:} Coping with the Illness of A Family Member \\
    \textbf{Situation:} My sister was recently hospitalized after a serious accident, and I'm feeling overwhelmed by the uncertainty and fear for her recovery. \\
    \midrule
    \midrule   
    \textbf{ASSISTANT:} \textit{``It's completely understandable to feel overwhelmed and scared when someone close to you is going through such a difficult time. Right now, it might be helpful to focus on what actions you can take to support your sister and yourself. For example, staying informed about her condition and communicating regularly with the medical team.''} \\

    \textbf{USER:}  \textit{``I think I could start by talking to the doctors more often to understand her condition better and reaching out to a close friend for emotional support.''} \\

    \textbf{ASSISTANT:} \textit{``That's a great start! Talking to the doctors more frequently can provide you with clearer insights into your sister's condition. Additionally, sharing your concerns and experiences with someone you trust can make a significant difference in how you cope with this situation. Remember, it's okay to lean on others during tough times. How do you feel about taking these steps?''} \\

    \textbf{USER:}  \textit{``I feel a bit nervous about opening up to others, but I know I need to try, and I’m worried about staying strong for my sister while managing my own emotions.''} \\
    
    \textbf{ASSISTANT:} \textit{``It's completely normal to feel nervous about opening up, especially when you're already dealing with so much. However, sharing your feelings with someone you trust can actually help you stay strong in the long run. As for being there for your sister, it's important to remember that you can't pour from an empty cup. Taking care of your own emotional well-being is not only necessary but also essential for supporting her. Maybe you could set boundaries or let her know that you're doing your best to handle everything.''} \\

    \textbf{USER:}  \textit{`` I think I could set small moments aside each day to focus on myself, like taking a walk or journaling, while letting my sister know I’m always there for her when she needs me.''} \\

    \textbf{ASSISTANT:} \textit{``That sounds like a very practical approach. Setting aside specific times for self-care, such as walking or journaling, can help you recharge and maintain your emotional strength. And letting your sister know that you're always there for her when she needs you can also help her feel supported and less alone. It's important to strike a balance between caring for yourself and being there for her. If you find that you're struggling to maintain this balance, don't hesitate to reach out for additional support. You're doing great by thinking ahead and planning ways to take care of yourself and your sister.''} \\

    \textbf{USER:}  \textit{``Thank you, I’ll try to take it one step at a time and remind myself that it’s okay to ask for help when I need it.''} \\
    \midrule
    \textbf{Reward: }0.55  $\rightarrow$  Goal completed !!\\

    \bottomrule
    \end{tabular}
    \caption{An example conversation with our RLFF-ESC on the ExTES dataset.}
    \label{tab:eample_2}
\end{table*}

\end{document}